\PassOptionsToPackage{table}{xcolor}
\documentclass[11pt]{article}

\usepackage{acl}
\usepackage{times}
\usepackage{latexsym}
\usepackage[T1]{fontenc}
\usepackage[utf8]{inputenc}
\usepackage{microtype}
\usepackage{inconsolata}
\usepackage{graphicx}

\usepackage{booktabs}
\usepackage{amsmath}
\usepackage{amssymb}
\usepackage{multirow}
\usepackage{xcolor}
\definecolor{rankfirst}{HTML}{2E7D32}
\definecolor{ranksecond}{HTML}{1565C0}
\definecolor{rankthird}{HTML}{C62828}
\usepackage{enumitem}
\usepackage{float}
\newcommand{\model}{N2NSC}

\title{\raisebox{-0.5ex}{\includegraphics[height=1.2em]{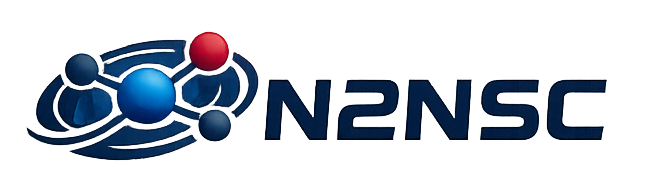}}~Node-to-Neighborhood Semantic Consistency:\\
Text-Topology Alignment for TAGs Anomaly Detection}

\author{
  Bochen Lin$^{1}$,
  Jianxiang Yu$^{1}$,
  Jiayi Wu$^{1}$,
  Lin Qi$^{2}$,
  Huang Lu$^{2}$,
  Xiang Li$^{1,*}$ \\
  $^{1}$School of Data Science and Engineering,
  East China Normal University, Shanghai, China \\
  $^{2}$WeChat, Tencent, Guangzhou, China \\
  \url{bochenlin@stu.ecnu.edu.cn}
}

\begin{document}
\maketitle
\renewcommand{\thefootnote}{$\ast$}
\footnotetext{Corresponding author.}
\setcounter{footnote}{3}
\renewcommand{\thefootnote}{\arabic{footnote}}
\footnotetext{\url{https://github.com/aibert2/N2NSC}}
\setcounter{footnote}{3}

\begin{abstract}

Graph anomaly detection (GAD) on text-attributed graphs (TAGs) is
vital for applications such as fraud detection and academic integrity
verification. Existing approaches generally fall into two paradigms.
GNN-based methods effectively capture structural patterns but struggle
to capture fine-grained textual semantics. Methods integrating LLMs
with graphs improve semantic understanding yet fail to fully comprehend
topological relationships among neighboring nodes. Moreover, both
paradigms overlook the correspondence between textual semantics and
graph topological relationships, limiting their ability to identify
nodes whose semantics are inconsistent with their neighborhoods. In
this paper, we formalize TAG anomaly detection as a
node-to-neighborhood semantic consistency problem, where anomalies may
arise from either textual semantic mismatch or topological deviation between a node and its neighbors. We propose \textsc{\model}{} (Node-to-Neighborhood Semantic Consistency), a framework that captures the correspondence between graph topology and textual semantics through two complementary fusion paths. The two pathways work synergistically, enabling the LLM to fully leverage both textual and structural neighborhood
information for anomaly detection. Extensive experiments across eight
datasets demonstrate that \textsc{\model}{} consistently
outperforms current state-of-the-art methods. Our code is available.\footnotemark[3]

\end{abstract}

\begin{figure*}[t]
  \centering
  \includegraphics[width=\textwidth]{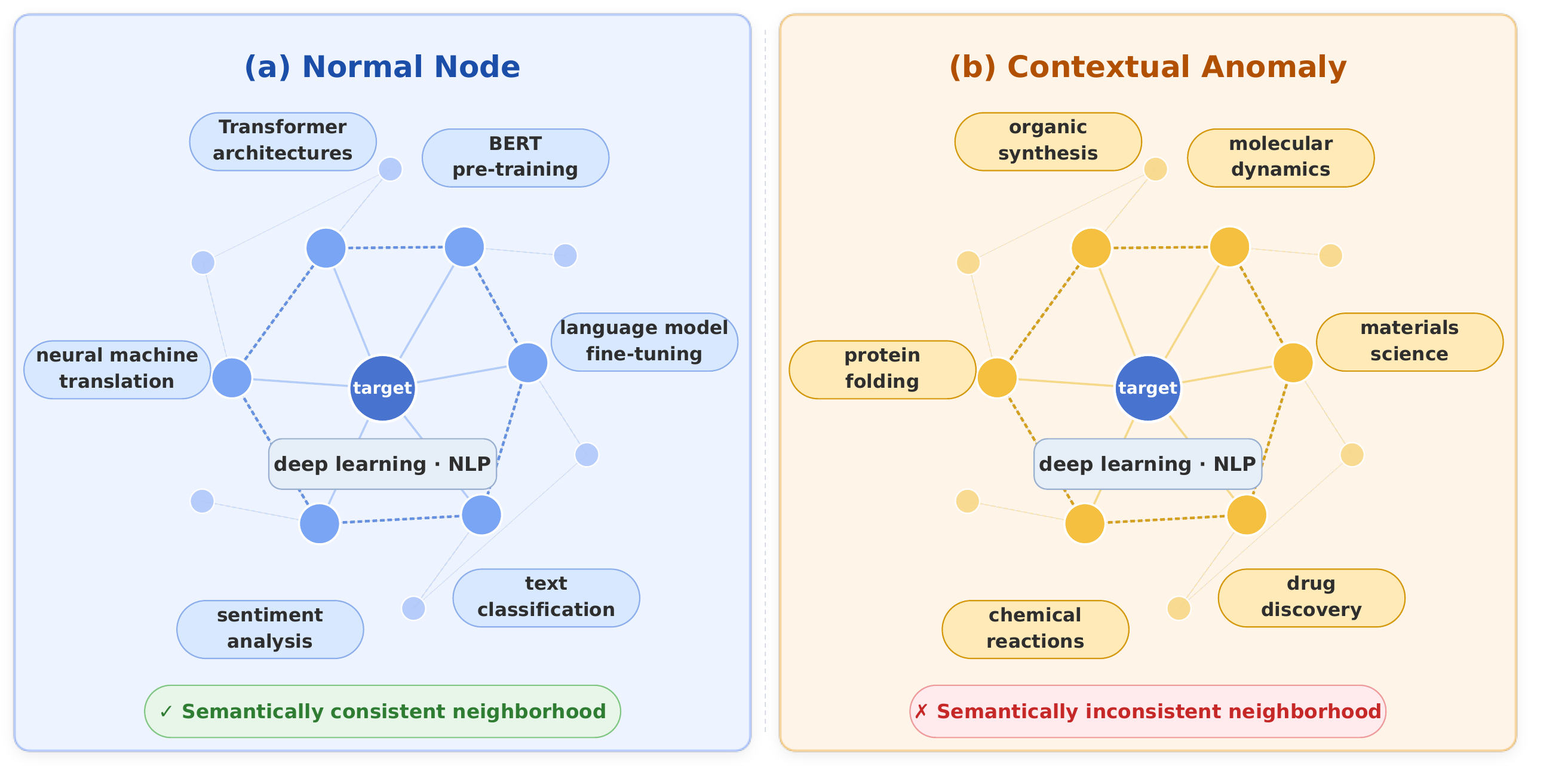}
  \caption{
    \textbf{Neighborhood semantic inconsistency in text-attributed graphs.}
    Both panels show the same target node (``deep learning $\cdot$ NLP'').
    In~(a), every neighbor belongs to the NLP domain, and the neighborhood is
    semantically consistent.
    In~(b), all neighbors are from chemistry, and the neighborhood is
    semantically inconsistent, making the node anomalous due to
    \emph{node-to-neighborhood semantic inconsistency}.
    The anomaly signal is relational, not lexical: the target text is
    identical in both cases.
  }
  \label{fig:motivation}
\end{figure*}

\section{Introduction}

Anomaly detection on graphs has become increasingly important in
graph mining~\citep{chandola2009anomaly,ding2019deep,liu2021pick,tang2022rethinking,xu2025cmucl},
driven by its broad applicability to fraud detection and scientific integrity analysis. While substantial progress has been
made on attributed graphs with fixed-dimensional node
features~\citep{liu2021pick,roy2024gadnr,ju2025survey}, real-world networks
increasingly exhibit richer representations~\citep{yan2023tag}. In citation
networks~\citep{hu2020ogb}, nodes carry paper abstracts. In e-commerce
platforms, product descriptions serve as node attributes. For anomaly
detection on such text-attributed graphs (TAGs)~\citep{su2025llmtagssurvey}, the anomaly signal
depends not only on a node's own textual content, but also on the
textual content of its neighbors and the topological relationships
among them.

We refer to this phenomenon as \emph{node-to-neighborhood semantic inconsistency}~\citep{song2007conditional}, where a node’s semantics are inconsistent with the semantic and topological context of its neighborhood.
Such inconsistency may manifest as textual semantic deviation, where
a node's text diverges from what its neighbors discuss, or as
structural topological anomaly, where a node's connectivity pattern
deviates from its local graph context.
This is illustrated in Figure~\ref{fig:motivation}. Consider a paper on deep learning for NLP. If all its citing neighbors are also NLP papers, the neighborhood is semantically consistent and the node is normal. If instead the same paper is surrounded by chemistry publications, the semantic mismatch with its neighborhood makes it anomalous. The node 
text remains identical in both cases. The anomaly signal arises from
node-to-neighborhood semantic inconsistency rather than any intrinsic
textual defect.

Existing methods have yet to address this problem adequately. GNN-based methods~\citep{tang2022rethinking,liu2021pick,dou2020care,roy2024gadnr,ma2022deep} model graph topology through message passing but reduce text to frozen embedding vectors, making fine-grained textual reasoning unavailable at decision time. Methods integrating LLMs with graphs~\citep{chen2024llaga,tang2024graphgpt,li2024glbench} bring language understanding to graph tasks, yet existing approaches either use LLMs as feature generators~\citep{he2024harnessing}, jointly train text and graph encoders with contrastive objectives~\citep{xu2025cmucl}, or employ multi-LLM collaboration for evidence aggregation~\citep{xu2025coll}. In these approaches, the model either has no access to what neighboring nodes actually say, has no awareness of graph-topological relationships, or receives both forms of information yet lacks the capacity to reason over them jointly.\looseness=-1

In this work, we introduce \textsc{\model} (\emph{Node-to-Neighborhood
Semantic Consistency}), a framework that addresses this limitation
through two complementary fusion paths, each capturing the correspondence between graph topology and textual semantics through a different mechanism.
The \textbf{explicit fusion path} directly shapes what the LLM reads
by fusing textual neighborhood semantics and graph-structural
neighborhood semantics into a unified representation, enabling the LLM to
reason over the full graph-text context of each node.
The \textbf{implicit fusion path} transforms neighborhood semantics into node-adaptive modulation
signals that guide the language model to understand the correspondence
between graph topology and textual semantics for each node.
Together, the two paths enable the language model to fully leverage
both textual and structural neighborhood information for anomaly
detection.\looseness=-1

Extensive experiments on eight benchmark TAG anomaly detection
datasets~\citep{xu2025cmucl}, spanning citation networks, e-commerce
networks, and large-scale bibliographic data, show that \textsc{\model}
consistently outperforms state-of-the-art GAD and LLM-based baselines.
Key contributions of this work are as follows.
\begin{itemize}[nosep,leftmargin=*]
\item We formalize TAG anomaly detection as a node-to-neighborhood
  semantic consistency problem and propose \textsc{\model}, a framework
  that deeply integrates graph-structural semantics and textual
  semantics into LLMs through two complementary fusion paths.
\item We design an explicit fusion path that integrates textual and topological neighborhood semantics into a unified representation, alongside an implicit fusion path that transforms neighborhood semantics into node-adaptive parameter modulation, enabling the LLM to dynamically adapt its computation to each node’s graph context.
\item Extensive experiments on eight benchmark datasets validate that
  \textsc{\model} achieves consistent and significant improvements
  over 17 baselines spanning both GNN-based and LLM-integrated methods.
\end{itemize}

\vspace{-0.2\baselineskip}
\section{Related Work}

\paragraph{Graph anomaly detection.}

Graph anomaly detection has attracted broad research
interest~\citep{ma2026lect,xu2024llmanomaly}.
Existing methods generally fall into two paradigms.
The first relies on GNN-based classifiers operating on fixed node
embeddings.
Representative approaches include reconstruction-based autoencoders
such as DOMINANT~\citep{ding2019deep}, spectral-wavelet filters
(BWGNN~\citep{tang2022rethinking}), imbalance-aware neighbor sampling
(PC-GNN~\citep{liu2021pick}), camouflage-resistant relational learning
(CARE-GNN~\citep{dou2020care}), and neighborhood-reconstruction
objectives (GAD-NR~\citep{roy2024gadnr}).
Contrastive self-supervised methods such as
CoLA~\citep{liu2021cola}, ANEMONE~\citep{jin2021anemone}, and
GCCAD~\citep{chen2022gccad} further enrich structural representations.
These methods treat text solely as frozen pretrained embedding vectors and therefore fail to capture fine-grained textual semantic patterns.
The second paradigm attempts to integrate textual semantics with graph structure through language models.
TAPE~\citep{he2024harnessing} uses LLM-generated explanations as
enhanced node features for downstream classification.
CMU-CL~\citep{xu2025cmucl} jointly trains text and graph encoders with
contrastive consistency.
GuARD~\citep{pang2025guard} combines GNN embeddings with LM features
through multi-modal instruction tuning.
CoLL~\citep{xu2025coll} employs multi-LLM collaboration for
evidence-augmented generation.
In a related direction, UNPrompt~\citep{niu2025unprompt} learns unified
graph prompts from neighborhood structure for zero-shot anomaly
detection without relying on language models.
TAGAD~\citep{liu2025towards} detects contextual anomalies by comparing ego-graph and text-graph representations, yet its dual modules operate independently without jointly reasoning over both modalities. Overall, these methods treat graph information as auxiliary signals rather than fully exploiting the correspondence between graph topology and textual semantics.

\paragraph{Graph-text semantic fusion in language models.}
Existing approaches to integrating graph-structural semantics into language models~\citep{li2024glbench,xu2024llmanomaly} follow two strategies. Some inject graph signals into the model input as additional text or tokens, such as structural descriptions~\citep{ye2024instructglm} or LLM-generated explanations~\citep{he2024harnessing}. Others align graph and text representations in the embedding space through contrastive objectives~\citep{xu2025cmucl} or cross-modal projection~\citep{tang2024graphgpt,chen2024llaga,wang2024bridging}.
All these approaches confine graph semantics to either the input or the embedding space, without capturing the correspondence between graph topology and textual semantics.
In contrast, \textsc{\model} achieves semantic fusion through two
complementary levels simultaneously.
The explicit path fuses textual neighborhood semantics and
graph-structural neighborhood semantics at the input level, while
the implicit path transforms neighborhood semantics into
node-adaptive modulation signals that guide the model to understand
the correspondence between graph topology and textual semantics.
This dual-path design provides a capability that no prior method achieves.

\section{Preliminaries}
\label{sec:preliminary}

Let $\mathcal{G} = (\mathcal{V}, \mathcal{E}, \mathbf{X}, \mathbf{T})$
denote a text-attributed graph (TAG), where $\mathcal{V}$ is the set of
nodes, $\mathcal{E} \subseteq \mathcal{V} \times \mathcal{V}$ is the set
of edges, $\mathbf{X} \in \mathbb{R}^{|\mathcal{V}| \times d}$ is the
node feature matrix with each row $\mathbf{x}_v \in \mathbb{R}^d$ being
the feature vector of node $v$ obtained from its text via a pretrained
language encoder, and $\mathbf{T} = \{T_v\}_{v \in \mathcal{V}}$ is the
collection of raw node texts.
We denote the adjacency matrix as
$\mathbf{A} \in \{0,1\}^{|\mathcal{V}| \times |\mathcal{V}|}$, where
each entry $\mathbf{A}_{vu}$ indicates the link between nodes $v$ and
$u$.
Each node $v \in \mathcal{V}$ is associated with a binary label
$y_v \in \{0,1\}$ indicating whether the node is normal or anomalous.
Given a set of labeled nodes
$\mathcal{V}_L \subset \mathcal{V}$, the goal is to accurately identify
anomalous nodes in
$\mathcal{V}_U = \mathcal{V} \setminus \mathcal{V}_L$.

\section{Methodology}
\label{sec:method}

\subsection{Overview}

\begin{figure*}[t]
  \centering
  \includegraphics[width=\textwidth]{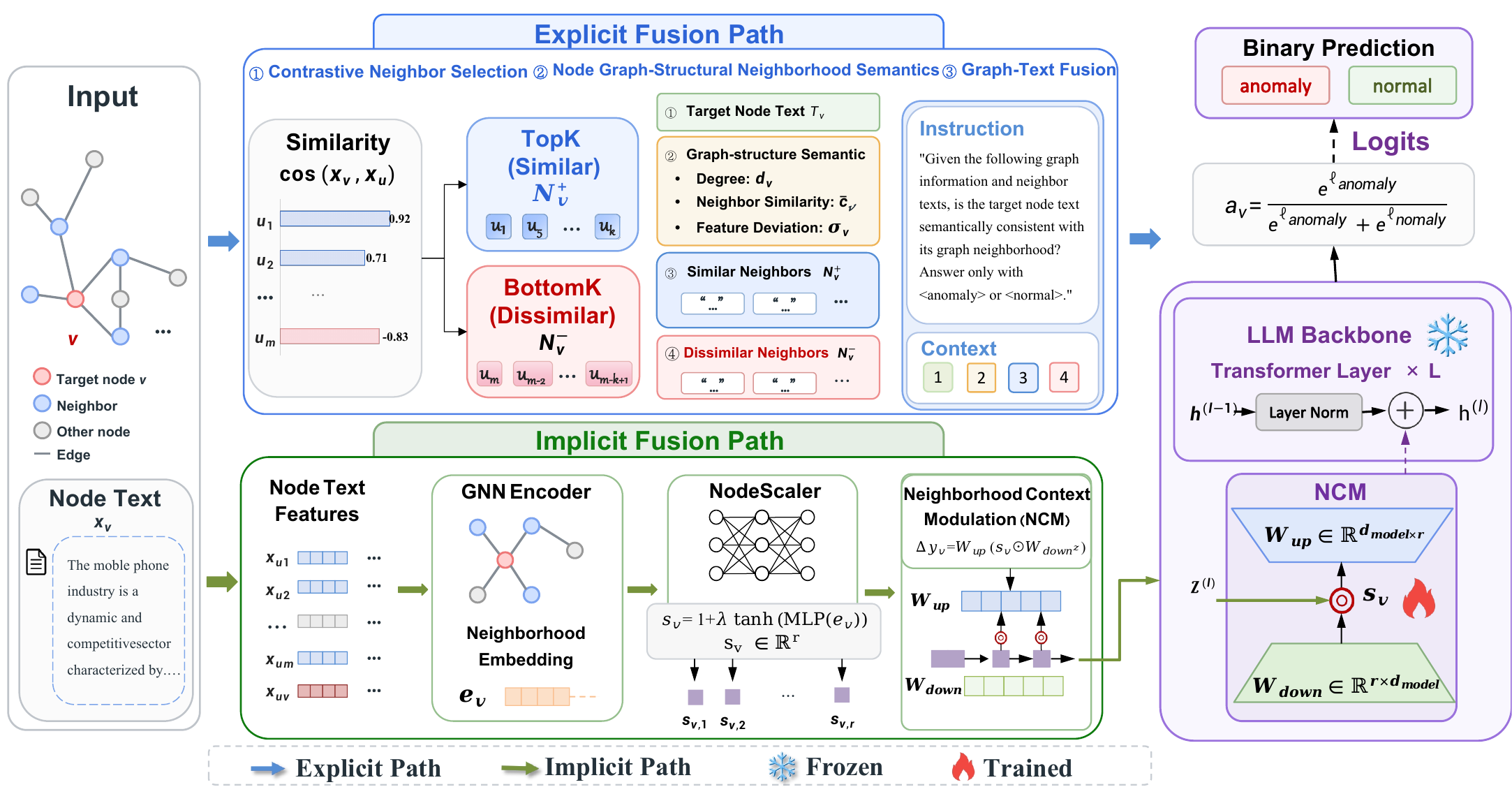}
  \caption{\textsc{\model} framework overview. The explicit fusion path
    fuses textual neighborhood semantics and graph-structural neighborhood
    semantics into a unified representation, while the implicit fusion path
    (NCM) transforms neighborhood semantics into node-adaptive
    modulation signals that guide the LLM to understand the correspondence between graph topology and textual semantics.}
  \label{fig:architecture}
\end{figure*}

Detecting node-to-neighborhood semantic inconsistency requires deeply fusing graph-structural semantics with textual semantics. \textsc{\model} achieves this through two complementary fusion paths that operate synergistically. The \textbf{explicit fusion path} fuses textual neighborhood semantics and graph-structural neighborhood semantics into a unified representation, providing the LLM with complete neighborhood semantics.
The \textbf{implicit fusion path} transforms neighborhood semantics into node-adaptive modulation signals that guide the language
model to understand the correspondence between graph topology and textual semantics for each node. Figure~\ref{fig:architecture} illustrates the framework.

\subsection{Explicit Fusion Path}
\label{sec:gsd}

The explicit path fuses graph-structural neighborhood semantics and
textual neighborhood semantics into a unified representation. It
captures the node's topological profile through neighborhood-derived
statistics, namely degree, neighbor similarity, and feature deviation,
along with contrastive neighbor texts from the node's graph
neighborhood $\mathcal{N}(v) = \{u : \mathbf{A}_{vu} = 1\}$.

\paragraph{Textual Neighborhood Semantics.}
For each target node $v$, we compute pairwise cosine similarity
between $\mathbf{x}_v$ and all graph neighbors:
\begin{equation}
  c_{vu} = \cos(\mathbf{x}_v,\, \mathbf{x}_u), \quad u \in \mathcal{N}(v).
  \label{eq:cosine}
\end{equation}
Two contrastive sets are selected.
$\mathcal{N}^{+}_v = \operatorname{TopK}_u\, c_{vu}$ contains
semantically similar neighbors, forming a \emph{local prototype}, and
$\mathcal{N}^{-}_v = \operatorname{BottomK}_u\, c_{vu}$ contains
semantically dissimilar neighbors, exposing \emph{local conflicts}.

\paragraph{Graph-Structural Neighborhood Semantics.}
We characterize the node's topological context through three
neighborhood-derived statistics.
Let $\bar{\mathbf{x}}_v = \frac{1}{|\mathcal{N}(v)|}
\sum_{u \in \mathcal{N}(v)} \mathbf{x}_u$ denote the mean feature
vector of node $v$'s neighbors.
We define the degree $d_v = |\mathcal{N}(v)|$, the average neighbor
similarity
$\bar{c}_v = \cos(\mathbf{x}_v,\, \bar{\mathbf{x}}_v)$,
and the feature deviation
$\delta_v = \|\mathbf{x}_v - \bar{\mathbf{x}}_v\|_2$.
A normal node typically has high $\bar{c}_v$ and low $\delta_v$,
while an anomalous node exhibits the opposite pattern.

\paragraph{Semantic Fusion.}
The fused representation is assembled into a structured input
comprising the raw text $T_v$ of the target node, the
neighborhood-derived statistics $d_v$, $\bar{c}_v$, $\delta_v$,
and extractive text snippets from $\mathcal{N}^{+}_v$ and
$\mathcal{N}^{-}_v$.
The LLM is then asked a binary consistency question: \emph{is the node
text semantically consistent with its graph neighborhood?}

By jointly presenting graph-structural neighborhood semantics and
textual neighborhood semantics in a unified representation, the
explicit path enables the LLM to reason over the full graph-text
context of each node without relying on hand-crafted rules
or LLM-generated summaries.

\subsection{Implicit Fusion Path}
\label{sec:ncm}

The implicit path fuses graph-text semantics into the LLM's parameter
space through \textbf{Neighborhood Context Modulation (NCM)}. The core
idea is to transform each node's neighborhood semantics into a
node-adaptive modulation signal that guides the language model to
understand the correspondence between graph topology and textual semantics for each node.

Concretely, a GNN encoder first aggregates each node's textual feature
$\mathbf{x}_v$ over its neighborhood structure via multi-hop message
passing, producing a \emph{neighborhood embedding} that jointly encodes
text semantics and graph context. A learned NodeScaler then
maps this embedding to a per-node scaling vector:
\begin{equation}
  \mathbf{s}_v = \mathbf{1} + \lambda\tanh(\mathrm{MLP}(\mathrm{GNN}(\mathbf{x}_v))),
  \label{eq:nodescaler}
\end{equation}
where $\lambda$ controls the maximum deviation from identity and
$\mathbf{1}$ denotes the all-ones vector.
This scaling vector modulates the intermediate representation of a
low-rank parameter update $\Delta W = W_{\mathrm{up}} W_{\mathrm{down}}$
applied to the frozen LLM weights:
\begin{equation}
  \Delta \mathbf{y}_v = W_{\mathrm{up}}(\mathbf{s}_v \odot W_{\mathrm{down}}\mathbf{z}),
  \label{eq:ncm}
\end{equation}
where $\mathbf{z}$ is the layer's hidden state, $\Delta \mathbf{y}_v$
is the modulated output for node $v$, and $\odot$ denotes
elementwise multiplication.
At initialization, $\mathbf{s}_v = \mathbf{1}$, which recovers the standard
parameter update. During training, NodeScaler learns to selectively
amplify or suppress individual channels per node, creating a complementary path through which neighborhood semantics guide the LLM to understand the correspondence between graph topology and textual semantics beyond what the explicit input alone conveys.

\subsection{Training and Inference}

The LLM backbone is frozen.
Only the low-rank weights $\{W_{\mathrm{up}}, W_{\mathrm{down}}\}$ and
NodeScaler parameters are trainable, optimized jointly via
cross-entropy over two answer tokens
(\texttt{anomaly}, \texttt{normal}).
At inference, the score for node $v$ is:
\begin{equation}
  a_v = \frac{\exp(\ell_{\texttt{anomaly}})}{\exp(\ell_{\texttt{anomaly}}) + \exp(\ell_{\texttt{normal}})},
  \label{eq:score}
\end{equation}
where $\ell_{\texttt{anomaly}}$ and $\ell_{\texttt{normal}}$ are the
logits of the two answer tokens given the assembled input
$\mathrm{ctx}_v$, which comprises the node text $T_v$,
graph-structural neighborhood semantics, and contrastive neighbor
texts as described above.

All input information is constructed without validation or test labels. The GNN encoder trains only on labeled nodes $\mathcal{V}_L$. The explicit path uses only graph topology, node features, and raw text.

\section{Experiments}
\label{sec:exp}

\subsection{Datasets}

We evaluate on eight text-attributed graph anomaly detection benchmarks~\citep{xu2025cmucl}: Citeseer, Pubmed, History, Photo, Computers, Children, ogbn-Arxiv, and CitationV8. The datasets cover both citation and e-commerce graphs, spanning diverse text domains and graph scales from 3K to over 1M nodes. Full statistics are given in Table~\ref{tab:datasets}.

\begin{table}[t]
\centering\small\setlength{\tabcolsep}{4pt}
\resizebox{\columnwidth}{!}{%
\begin{tabular}{lrrrrr}
\toprule
Dataset & Nodes & Edges & Avg.\ len. & Anomalies & Ratio (\%) \\
\midrule
Citeseer   &     3,186 &     3,432 & 153.94 &     128 & 4.02 \\
Pubmed     &    19,717 &    90,368 & 256.08 &     788 & 4.00 \\
History    &    41,551 &   369,252 & 228.36 &   1,662 & 4.00 \\
Photo      &    48,362 &   512,933 & 150.25 &   1,934 & 4.00 \\
Computers  &    87,229 &   742,792 &  93.16 &   3,490 & 4.00 \\
Children   &    76,875 & 1,574,664 & 209.12 &   3,076 & 4.00 \\
ogbn-Arxiv &   169,343 & 1,210,112 & 179.70 &   6,774 & 4.00 \\
CitationV8 & 1,106,759 & 6,396,265 & 148.77 &  44,270 & 4.00 \\
\bottomrule
\end{tabular}%
}
\caption{Dataset statistics. Avg.\ len.\ =
  whitespace-tokenised document length. All attributes are 768-dim.\
  frozen BGE embeddings. Anomaly ratio fixed at $\approx$4\%
  (50\% attribute-perturbed + 50\% structure-perturbed, no overlap).}
\label{tab:datasets}
\end{table}

\begin{table*}[!t]
\centering
\small
\setlength{\tabcolsep}{6pt}
\resizebox{\textwidth}{!}{%
\begin{tabular}{lrrrrrrrrr}
\toprule
Method & Citeseer & Pubmed & History & Photo & Computers & Children & ogbn-Arxiv & CitationV8 & Avg.\ Rank \\
\midrule
\multicolumn{10}{l}{\textit{GNN-based methods}} \\
GCN
  & 55.81 & 45.32 & 54.58 & 48.31 & 57.19 & 38.36 & 59.90 & 62.20 & 7.25 \\
GAT
  & 34.48 & 35.41 & 46.63 & 48.21 & 52.47 & 45.49 & 59.59 & 48.31 & 9.62 \\
GIN
  & 58.54 & 33.68 & 51.93 & 54.47 & \textcolor{rankthird}{\textbf{62.86}} & \textcolor{ranksecond}{\textbf{54.31}} & 42.90 & 62.96 & 7.00 \\
GraphSAGE
  & \textcolor{ranksecond}{\textbf{63.70}} & \textcolor{rankthird}{\textbf{72.37}} & 43.46 & 39.53 & 48.71 & 34.29 & 92.12 & 84.38 & \textcolor{rankthird}{\textbf{5.50}} \\
PC-GNN
  & 42.42 & 61.99 & 45.88 & 31.74 & 45.63 & 30.90 & 90.39 & 82.27 & 7.25 \\
BWGNN
  & \textcolor{rankthird}{\textbf{61.54}} & 47.11 & 21.96 & 18.52 & 34.71 & 19.90 & \textcolor{ranksecond}{\textbf{94.52}} & \textcolor{ranksecond}{\textbf{85.16}} & 8.00 \\
GHRN
  & 38.89 & 43.12 & 26.04 & 20.02 & 33.66 & 19.94 & \textcolor{rankthird}{\textbf{92.97}} & \textcolor{rankthird}{\textbf{84.28}} & 9.25 \\
XGBGraph
  & 52.89 & 59.44 & \textcolor{ranksecond}{\textbf{60.88}} & \textcolor{rankthird}{\textbf{48.97}} & \textcolor{ranksecond}{\textbf{57.34}} & \textcolor{rankthird}{\textbf{38.18}} & 89.99 & 72.24 & \textcolor{ranksecond}{\textbf{5.00}} \\
RFGraph
  & 47.06 & 42.09 & \textcolor{rankthird}{\textbf{59.64}} & 45.60 & 54.44 & 34.17 & 78.35 & 63.24 & 7.62 \\
GAAP
  & 18.95 & 51.80 & 19.31 & 17.72 & 25.50 & 16.17 & 79.23 & 61.86 & 12.25 \\
PMP
  & 19.44 & 58.27 & 21.06 & 21.33 & 24.89 & 11.81 & 73.45 & 49.63 & 12.38 \\
\midrule
\multicolumn{10}{l}{\textit{Integrating LLMs with graphs}} \\
TAPE
  & 58.54 & \textcolor{ranksecond}{\textbf{76.97}} & 49.68 & \textcolor{ranksecond}{\textbf{57.82}} & 53.44 & 48.36 & 65.28 & 68.21 & 5.25 \\
CoLL
  & 25.07 & 19.71 & 25.70 & 16.85 & 22.91 & 15.56 & 36.58 & 47.88 & 14.62 \\
UNPrompt
  & 23.18 & 16.14 & 26.73 & 11.86 & 14.30 & 11.82 & 37.65 & 35.63 & 15.12 \\
GraphGPT
  & 14.78 & 11.52 & 8.96 & 8.59 & 10.66 & 8.65 & 16.81 & 15.03 & 17.75 \\
InstructGLM
  & 9.21 & \textcolor{rankthird}{\textbf{61.34}} & 12.95 & 21.70 & 27.51 & 12.91 & 64.60 & 63.81 & 12.00 \\
ConsisGAD
  & 13.48 & 43.88 & 17.10 & 21.11 & 26.94 & 9.93 & 73.40 & 17.28 & 14.12 \\
\midrule
\multicolumn{10}{l}{\textit{Ours}} \\
\rowcolor{gray!15}
\textbf{\textsc{\model}}
  & \textcolor{rankfirst}{\textbf{87.50}} & \textcolor{rankfirst}{\textbf{93.46}} & \textcolor{rankfirst}{\textbf{84.70}} & \textcolor{rankfirst}{\textbf{66.76}} & \textcolor{rankfirst}{\textbf{85.78}} & \textcolor{rankfirst}{\textbf{70.40}} & \textcolor{rankfirst}{\textbf{98.35}} & \textcolor{rankfirst}{\textbf{98.53}} & \textcolor{rankfirst}{\textbf{1.00}} \\
Improve.
  & +23.80 & +16.49 & +23.82 & +8.94 & +22.92 & +16.09 & +3.83 & +13.37 & -- \\
\bottomrule
\end{tabular}%
}

\caption{Anomaly-F1 (\%) on eight TAG benchmarks.
  Performance comparison of GNN-based methods and LLM-graph methods.
  \textcolor{rankfirst}{\textbf{First}},
  \textcolor{ranksecond}{\textbf{second}}, and
  \textcolor{rankthird}{\textbf{third}} best results are color-coded.}
\label{tab:baseline}
\end{table*}
\subsection{Baselines}

To verify the effectiveness of our proposed method, we select several
baseline models for comparison, categorized into two types.

\noindent\textbf{GNN-based methods.}
GCN~\citep{kipf2016gcn},
GAT~\citep{velivckovic2018gat},
GIN~\citep{xu2018gin},
GraphSAGE~\citep{hamilton2017sage},
PC-GNN~\citep{liu2021pick},
BWGNN~\citep{tang2022rethinking},
GHRN~\citep{gao2023ghrn},
XGBGraph and RFGraph~\citep{chen2016xgboost},
GAAP~\citep{duan2025gaap}, and PMP~\citep{zhuo2024pmp}.
These methods classify on fixed embeddings without language model
reasoning.

\noindent\textbf{Integrating LLMs with graphs.}
TAPE~\citep{he2024harnessing},
CoLL~\citep{xu2025coll},
UNPrompt~\citep{niu2025unprompt},
GraphGPT~\citep{tang2024graphgpt},
InstructGLM~\citep{ye2024instructglm}, and
ConsisGAD~\citep{chen2024consisgad}.
These methods attempt to integrate language understanding with graph structure for node-level tasks.

\subsection{Experimental Setup}
We employ a stratified 60/20/20 split to partition the training,
validation, and test sets, and all experiments are conducted on this
shared partition.
For GNN-based methods, we utilize the 768-dim node features generated
by embedding the raw text of all nodes using LM-based BGE~\citep{xiao2024c}.
For methods that integrate LLMs with graph structures, we provide the
raw text of each node as input along with the same BGE embeddings and
graph topology.
Given that integrating LLMs for TAG anomaly detection remains an
emerging field, most LLM-graph baselines were originally designed for
other tasks (such as node classification or link prediction) or
different feature spaces (such as bag-of-words).
We adapt them to a unified evaluation protocol to provide the most
informative comparison possible (adaptation details in
Appendix~\ref{app:baselines}).

Regarding our proposed method, we select Qwen3-8B~\citep{yang2025qwen3} as the frozen LLM
backbone and employ LoRA~\citep{hu2022lora} with a rank of 64 for
parameter-efficient training. The GNN encoder adopts a two-layer GAT
architecture with 8 attention heads and an output dimension of 256.
In the contrastive neighbor selection phase, both TopK and BottomK
are set to $K{=}2$.

\subsection{Main Results}
\label{sec:main-comparison}

As shown in Table~\ref{tab:baseline}, \textsc{\model} ranks 1st on all
eight datasets with an average F1 of $85.69$, surpassing the best
baseline on each dataset by an average of $+16.16$ points. The largest
gains appear on History ($+23.82$) and Computers ($+22.92$), where
detecting anomalies requires jointly reasoning over textual and
structural neighborhood semantics. Even on ogbn-Arxiv, where
spectral-based GNN methods already achieve above $90$, \textsc{\model}
still improves by $+3.83$.

Both existing paradigms exhibit fundamental limitations when compared
with \textsc{\model}.
\textbf{(1) GNN-based methods show high instability across datasets}, with
average ranks ranging from $5.0$ to $12.4$. For example, BWGNN ranks
2nd on ogbn-Arxiv but drops to 13th on History, while XGBGraph ranks
2nd on History but only 6th on ogbn-Arxiv. This instability stems from
their reliance on static embeddings, which discard the fine-grained
textual cues needed for assessing node-to-neighborhood semantic
consistency.
\textbf{(2) Methods integrating LLMs perform even worse overall.} Despite having access to language understanding, these
methods inject graph information only as shallow auxiliary signals,
failing to capture the correspondence between graph topology and
textual semantics.
\textbf{(3)} In contrast, \textbf{\textsc{\model} achieves stable rank-1 performance}
through deep fusion of both modalities via the explicit and implicit
paths.

\begin{figure*}[t]
  \centering
  \includegraphics[width=\textwidth]{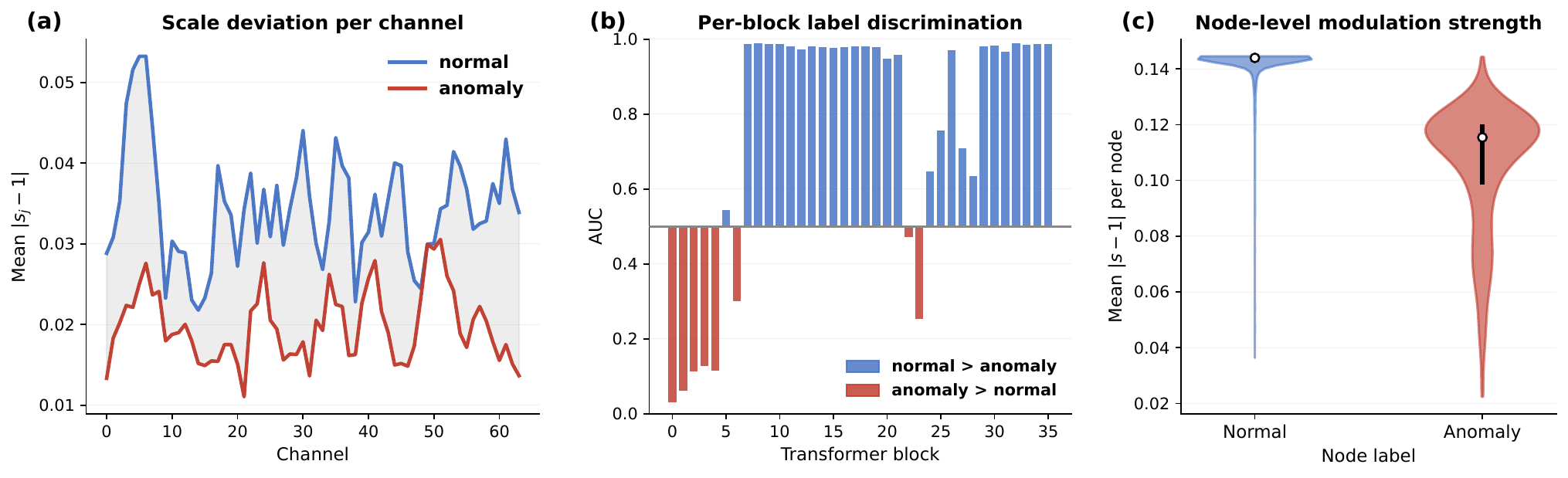}
  \caption{Mechanistic analysis of NCM on ogbn-Arxiv.
    \textbf{(a)} Per-channel mean $|s_j{-}1|$ averaged over the top-5
    most discriminative layers. Normal nodes receive consistently different
    modulation from anomaly nodes.
    \textbf{(b)} Per-block discriminative power. Bars extend upward from
    the 0.5 midline for blocks where normal nodes have larger deviation,
    downward where anomaly nodes do.
    \textbf{(c)} Violin plot of per-node mean $|s{-}1|$, showing that
    normal nodes receive systematically stronger modulation than anomaly
    nodes.}
  \label{fig:nodescaler}
\end{figure*}

\subsection{Ablation}

We validate each component of \textsc{\model} by progressively adding
them and observing the effect. Table~\ref{tab:ablation-feat} reports
results on Citeseer and Pubmed, and Table~\ref{tab:ablation-full}
reports full eight-dataset results.

\paragraph{Problem. Text alone cannot reveal neighborhood inconsistency.}
A text-only LLM achieves only $70.15$ AUROC on Citeseer and $73.78$ on
Pubmed, because anomalous nodes are textually plausible in isolation.
Adding neighborhood-derived statistics lifts AUROC to $92.74$ and
$97.91$, respectively, demonstrating that graph-structural context is
essential for detecting node-to-neighborhood semantic inconsistency.

\paragraph{Cause 1. Contrastive neighbor text is effective.}
Neighborhood statistics tell the LLM \emph{that} something is
inconsistent, but not \emph{what} the inconsistency is. Adding
contrastive neighbor text enables direct semantic comparison. To confirm
this, we compare against two controls where the neighbor text is
replaced with randomly sampled or shuffled content while keeping NCM
active. As shown in Table~\ref{tab:ablation-full}, both controls
degrade substantially, confirming that the quality of neighbor selection
matters. Furthermore, using only dissimilar neighbors outperforms using
only similar neighbors on most datasets, because dissimilar neighbors
directly expose the semantic conflicts that characterize anomalies.

\paragraph{Cause 2. NCM provides complementary gains.}
The explicit path captures what can be expressed in text, but some
anomaly patterns manifest as distributional shifts across the
neighborhood that text alone cannot convey. NCM addresses this by
transforming neighborhood semantics into node-adaptive modulation
signals. As shown in Tables~\ref{tab:ablation-auroc}
and~\ref{tab:ablation-auprc}, adding NCM consistently improves
threshold-free metrics across all eight datasets, with the largest
gains on large-scale graphs where distributional patterns are richer.
Figure~\ref{fig:nodescaler} provides mechanistic confirmation on
ogbn-Arxiv: Panel~(a) shows that normal nodes receive consistently
larger per-channel scaling deviations than anomaly nodes, Panel~(b)
reveals that the discriminative power concentrates in mid-to-late
Transformer blocks, and Panel~(c) confirms the separation at the
per-node level. A detailed analysis is provided in
Appendix~\ref{app:nodescaler-full}.

\begin{table*}[t]
\centering\small\setlength{\tabcolsep}{6pt}
\resizebox{\textwidth}{!}{%
\begin{tabular}{lcccccccc}
\toprule
 & & & & & \multicolumn{2}{c}{Citeseer} & \multicolumn{2}{c}{Pubmed} \\
\cmidrule(lr){6-7}\cmidrule(lr){8-9}
Variant & Deg & Cos & Feat & NCM & AUROC & AUPRC & AUROC & AUPRC \\
\midrule
Text only        & \texttimes & \texttimes & \texttimes & \texttimes & 70.15 & 44.55 & 73.78 & 52.94 \\
\midrule
Text + Degree    & \checkmark & \texttimes & \texttimes & \texttimes & 91.74 & 86.54 & 95.28 & 67.46 \\
Text + CosSim    & \texttimes & \checkmark & \texttimes & \texttimes & 90.94 & 71.24 & 84.71 & 58.81 \\
Text + FeatDev   & \texttimes & \texttimes & \checkmark & \texttimes & 65.83 & 44.96 & 80.03 & 53.87 \\
\midrule
Text + Deg\&Cos  & \checkmark & \checkmark & \texttimes & \texttimes & 90.62 & 85.52 & 98.20 & 86.79 \\
Text + All       & \checkmark & \checkmark & \checkmark & \texttimes & 92.74 & 85.93 & 97.91 & 86.70 \\
\rowcolor{gray!15}
Text + All + NCM & \checkmark & \checkmark & \checkmark & \checkmark & \textbf{93.28} & \textbf{86.06} & \textbf{98.44} & \textbf{88.19} \\
\bottomrule
\end{tabular}%
}
\caption{Feature ablation on Citeseer and Pubmed without contrastive
  neighbor text. All variants include node text as the base input.
  Deg = degree, Cos = neighbor similarity, Feat =
  feature deviation, NCM = Neighborhood Context Modulation.
  All values in \%.}
\label{tab:ablation-feat}
\end{table*}
\vspace{-1pt}

\subsection{Sensitivity Analysis}
\label{sec:sensitivity-analysis}

We analyze the sensitivity of \textsc{\model} along its two fusion paths
on Citeseer and Pubmed. All results are reported as
mean$\pm$std across the tested values of each hyperparameter.

\paragraph{Explicit fusion path. Neighbor count $K$.}
The explicit path selects the $K$ most similar and $K$ most dissimilar
neighbors as contrastive evidence. Across
$K \in \{1,2,3,4\}$, AUROC is $91.34{\pm}0.50$ on Citeseer and
$99.57{\pm}0.19$ on Pubmed, indicating that performance is highly
stable regardless of the number of selected neighbors.
$K{=}2$ achieves the best overall balance and is used as the default.

\paragraph{Implicit fusion path. Scaling strength $\lambda$.}
NCM modulates the parameter space via
$\mathbf{s}_v = \mathbf{1} + \lambda\tanh(\cdot)$, where $\lambda$
controls the maximum deviation from identity.
Across $\lambda \in \{0.1, 0.5, 1.0, 2.0\}$, AUROC is
$92.01{\pm}0.31$ on Citeseer and $99.62{\pm}0.09$ on Pubmed.
On Citeseer, all four values yield identical F1 of $89.80$.
This confirms that the residual initialization provides a stable
optimization landscape regardless of the maximum modulation magnitude.

\paragraph{Implicit fusion path. Modulation dimension $r$.}
The dimension $r$ determines the number of channels that NCM
modulates. Across $r \in \{16, 32, 64, 128\}$, performance on Pubmed is stable
with AUROC $99.57{\pm}0.11$ and F1 $93.27{\pm}1.07$.
On Citeseer, $r{=}128$ overfits with F1 dropping to $80.0$, consistent
with limited training data of approximately $1{,}900$ nodes.
Excluding this outlier, the remaining three values yield AUROC
$92.57{\pm}1.68$ and F1 $90.09{\pm}1.47$.
This motivates the default $r{=}64$ as a practical trade-off between
expressiveness and generalization across graph scales.

\subsection{Discussion}

\paragraph{Two fusion paths, one consistency objective.}
The ablation validates the two-path design. Graph-neighborhood semantic
cues yield $+22$ AUROC over text-only. Textual neighborhood semantics
outperform random and shuffled controls by $+4$ to $+15$. NCM adds
$+7$ to $+20$ AUPRC on large graphs. Removing any single component
causes degradation, confirming that the three components are
complementary. NCM is most valuable when explicit evidence is absent
or insufficient, reinforcing its role as a complementary fusion path.

\paragraph{Why extractive rather than aggregate?}
We compare against a variant that replaces neighbor text with a
statistical summary of the neighborhood. On Pubmed, the extractive
approach achieves $93.46$ F1 vs.\ $89.23$ for the aggregate variant,
because raw neighbor text preserves fine-grained semantic patterns
that summaries discard.

\paragraph{Distinction from retrieval-augmented generation.}
Standard RAG retrieves globally relevant documents to supply missing
knowledge. \textsc{\model}{} retrieves only from the target node's graph
neighborhood to assess node-to-neighborhood semantic consistency. The Random and
Shuffled controls confirm this distinction.

\paragraph{Scalability.}
\textsc{\model} scales to graphs with over one million nodes.
On CitationV8 (1.1M nodes, 6.4M edges), the framework achieves $98.53$
F1 with efficient training on a single GPU. The GNN encoder processes
neighborhoods locally, and the LLM evaluates one node at a time with
fixed-length input, keeping memory and computation manageable
regardless of overall graph size.

\paragraph{Generalization to real-world fraud.}
We further evaluate \textsc{\model} on two real-world fraud datasets, Amazon Video and YelpReviews~\citep{dou2020care}, where anomalies arise naturally
rather than through synthetic injection. As shown in Table~\ref{tab:real-fraud}, \textsc{\model} outperforms DGP~\citep{li2026dgp} by $+5$ to $+6$ AUROC and $+5$ to
$+7$ AUPRC, confirming that our framework generalizes to naturally occurring anomaly distributions.

\begin{table}[t]
\centering\small\setlength{\tabcolsep}{4pt}
\resizebox{\columnwidth}{!}{%
\begin{tabular}{llccc}
\toprule
Dataset & Method & AUROC & AUPRC & Macro-F1 \\
\midrule
\multirow{2}{*}{YelpReviews}
  & DGP           & 84.39 & 49.69 & 69.30 \\
  & \textbf{\textsc{\model}} & \textbf{89.47} & \textbf{55.34} & \textbf{71.19} \\
\midrule
\multirow{2}{*}{Amazon Video}
  & DGP           & 77.43 & 34.87 & 67.04 \\
  & \textbf{\textsc{\model}} & \textbf{83.16} & \textbf{41.52} & \textbf{73.28} \\
\bottomrule
\end{tabular}%
}
\caption{Results on real-world fraud datasets.}
\label{tab:real-fraud}
\end{table}
\vspace{-3pt}
\vspace{-3pt}

\label{page:mainmatter-end}

\section{Conclusion}

We formalize TAG anomaly detection as a node-to-neighborhood
semantic consistency problem and propose \textsc{\model}, a framework
that addresses it through two complementary fusion paths. The explicit
path fuses textual and graph-structural neighborhood semantics into a
unified representation. The implicit path transforms neighborhood
semantics into node-adaptive modulation signals that guide the language
model to understand the correspondence between graph topology and
textual semantics for each node. Meanwhile, our work reveals that capturing the correspondence between graph topology and textual semantics is essential for reasoning
over text-attributed graphs. Experiments across ten datasets validate that \textsc{\model} consistently outperforms state-of-the-art baselines.

\section*{Limitations}

This work introduces a framework that deeply fuses graph-structural
semantics with textual semantics for anomaly detection on
text-attributed graphs. However, it primarily focuses on the anomaly
detection task and cannot be directly applied to generative tasks such
as graph captioning, nor has it been evaluated on other discriminative
tasks like node classification. Additionally, the explicit fusion path
depends on the availability of neighborhood context, and its
effectiveness is reduced for isolated nodes with very few graph
neighbors. Leveraging this framework to construct a graph foundation
model presents a challenging yet valuable area for future exploration.

\section*{Ethics Considerations}

This work addresses graph anomaly detection for applications such as
fraud detection and academic integrity verification. All experiments
are conducted on publicly available benchmark datasets. The datasets
do not contain personally identifiable information, and no human
subjects are involved in this research. We acknowledge that anomaly
detection systems can be misused for surveillance or discriminatory
profiling.

\bibliography{custom}

\appendix

\section{Dataset Descriptions}
\label{app:datasets}

We evaluate on eight text-attributed graph anomaly detection benchmarks
introduced by~\citet{xu2025cmucl}, spanning citation networks
and Amazon e-commerce networks. All datasets share the same anomaly
injection protocol (4\% ratio, 50\% attribute-perturbed + 50\%
structure-perturbed, no overlap) and are represented as 768-dimensional
frozen BGE embeddings~\citep{xiao2024c}.

\paragraph{Citeseer.}
A citation network of 3,186 computer science publications connected by
3,432 citation links. Each node carries the paper title and abstract as
its text attribute, with an average document length of 154 tokens. The
graph is sparse (avg.\ degree $\approx$2.2), making neighborhood
evidence limited for low-degree nodes.

\paragraph{Pubmed.}
A biomedical citation network of 19,717 publications and 90,368 citation
edges. Node texts are paper titles and abstracts from the PubMed
database, averaging 256 tokens. The domain-specific vocabulary provides
strong lexical cues for neighborhood consistency.

\paragraph{History.}
An Amazon e-commerce graph of 41,551 products in the Books$\rightarrow$History
category, connected by 369,252 co-purchase edges. Node texts are book
titles and descriptions, averaging 228 tokens. The graph is denser than
citation networks (avg.\ degree $\approx$17.8).

\paragraph{Photo.}
An Amazon e-commerce graph of 48,362 products in the Electronics$\rightarrow$Photo
category with 512,933 co-purchase edges. Node texts are high-rated
reviews and product summaries, averaging 150 tokens.

\paragraph{Computers.}
An Amazon e-commerce graph of 87,229 products in the
Electronics$\rightarrow$Computers category with 742,792 co-purchase edges.
Node texts are high-rated reviews and product summaries, averaging 93
tokens (the shortest among all datasets).

\paragraph{Children.}
An Amazon e-commerce graph of 76,875 products in the Books$\rightarrow$Children
category with 1,574,664 co-purchase edges. Node texts are book titles
and descriptions, averaging 209 tokens. This is the densest graph in
our benchmark (avg.\ degree $\approx$41.0), providing rich neighborhood
context.

\paragraph{ogbn-Arxiv.}
A large-scale citation network of 169,343 computer science papers from
the Open Graph Benchmark~\citep{hu2020ogb} with 1,210,112 citation
edges. Node texts are paper titles and abstracts averaging 180 tokens.
The graph covers multiple CS sub-fields, creating natural topical
clusters.

\paragraph{CitationV8.}
The largest dataset in our benchmark, containing 1,106,759 papers and
6,396,265 citation edges from the DBLP Citation Network V8. Node texts
are paper titles and abstracts averaging 149 tokens. This dataset tests
scalability to million-node graphs.

\section{Baseline Descriptions}
\label{app:baselines}

We compare against 17 baseline methods from two families. Since the
the benchmark~\citep{xu2025cmucl} is recent and no prior work has
reported results on all eight datasets under identical conditions, we
adapt each baseline to the unified evaluation protocol (same 768-dim
BGE node features, same 60/20/20 split, same threshold calibration on
validation). For methods originally designed for different tasks or
feature spaces, we describe the adaptation below.

\subsection{GNN-Based Methods}

These methods classify on fixed 768-dim BGE node embeddings and train
on the same split as \textsc{\model}. We use official implementations
where available and tune hyperparameters on the validation set.

\paragraph{GCN~\citep{kipf2016gcn}.}
Graph Convolutional Network with symmetric normalized aggregation.
We use a two-layer architecture and tune hidden dimension on the
validation set.

\paragraph{GAT~\citep{velivckovic2018gat}.}
Graph Attention Network using multi-head attention for adaptive neighbor
weighting. Two-layer architecture with hyperparameters tuned per dataset.

\paragraph{GIN~\citep{xu2018gin}.}
Graph Isomorphism Network with sum aggregation and learnable epsilon.
Designed to be as expressive as the Weisfeiler-Leman graph isomorphism
test.

\paragraph{GraphSAGE~\citep{hamilton2017sage}.}
Inductive representation learning with mean aggregation and neighbor
sampling. Uses mini-batch training suitable for large graphs.

\paragraph{PC-GNN~\citep{liu2021pick}.}
Pick and Choose GNN designed for imbalanced graph classification.
Uses label-balanced neighbor sampling to address the severe class
imbalance (4\% anomaly ratio). We use the official implementation and
adapt it to BGE features.

\paragraph{BWGNN~\citep{tang2022rethinking}.}
Beta Wavelet GNN with spectrally localized band-pass filters.
Originally designed for lower-dimensional features. We adapt it to
768-dim BGE embeddings and tune hyperparameters per dataset.

\paragraph{GHRN~\citep{gao2023ghrn}.}
Graph Heterophily-aware Residual Network that prunes inter-class edges
by emphasizing high-frequency components. Adapted to the benchmark with the same features and split.

\paragraph{XGBGraph and RFGraph~\citep{chen2016xgboost}.}
Gradient-boosted trees (XGBoost) and Random Forest. We concatenate
node features with hand-crafted graph statistics (degree, clustering
coefficient, PageRank) as input, following the GADBench
protocol~\citep{tang2023gadbench}.

\paragraph{GAAP~\citep{duan2025gaap}.}
Global Attribute-Association Pattern aggregation for graph fraud
detection. Captures high-order attribute association patterns among
nodes for anomaly scoring. We adapt it to BGE features on the
benchmark split.

\paragraph{PMP~\citep{zhuo2024pmp}.}
Partitioning Message Passing for graph fraud detection. Partitions
the message-passing process to separate benign and fraudulent
propagation patterns. Adapted to the benchmark with the same features
and split.

\subsection{Methods Integrating LLMs with Graphs}

These methods were originally designed for different datasets or tasks.
We adapt them to the benchmark by providing the same BGE
features and graph topology under the unified split. Since TAG anomaly
detection with LLMs is a nascent area, most methods below are adapted
from related graph-LLM tasks rather than reproduced from native TAG-GAD
settings.

\paragraph{TAPE~\citep{he2024harnessing}.}
Originally proposed for node classification. Uses LLM-generated
explanations as enhanced node features for downstream GNN training.
We adapt it by replacing the classification objective with binary
anomaly detection on the benchmark split using official source code.

\paragraph{CoLL~\citep{xu2025coll}.}
Multi-LLM collaboration for TAG anomaly detection. This is one of the
few methods natively designed for this task. We use official source code
with the same benchmark data and split.

\paragraph{UNPrompt~\citep{niu2025unprompt}.}
Learns unified graph prompts for zero-shot GAD. Originally pre-trains a
GCN backbone on Facebook and transfers to target graphs. We follow the
original protocol (pre-train on Facebook, fine-tune prompts on the benchmark
train split). Note that UNPrompt reduces features to 8-dim via SVD,
discarding most of the 768-dim BGE information.

\paragraph{GraphGPT~\citep{tang2024graphgpt}.}
Graph instruction-tuning framework originally for node classification
and link prediction. We adapt it to anomaly detection by reformulating
the task as binary classification.

\paragraph{InstructGLM~\citep{ye2024instructglm}.}
Instruction-tuned graph language model that converts graph structures
into natural language descriptions. Adapted to the anomaly detection
task on the benchmark data.

\paragraph{ConsisGAD~\citep{chen2024consisgad}.}
Consistency-based graph anomaly detection that leverages consistency
between multiple augmented views of graph data for anomaly scoring.
We use the provided implementation on the benchmark features.

\section{Additional Results}
\label{app:more-results}

Tables~\ref{tab:auroc-full} and~\ref{tab:auprc-full} report
threshold-free AUROC and AUPRC for \textsc{\model} and all baseline
methods, complementing the Anomaly-F1 comparison in
Table~\ref{tab:baseline}.

\begin{table*}[!ht]
\centering
\footnotesize
\setlength{\tabcolsep}{4pt}
\resizebox{\textwidth}{!}{%
\begin{tabular}{lrrrrrrrrr}
\toprule
Method & Citeseer & Pubmed & History & Photo & Computers & Children & ogbn-Arxiv & CitationV8 & Avg.\ Rank \\
\midrule
\multicolumn{10}{l}{\textit{GNN-based methods}} \\
GCN          & \textcolor{rankthird}{\textbf{90.04}} & 91.61 & 78.20 & 78.48 & 81.81 & 75.65 & 95.00 & 90.01 & 7.38 \\
GAT          & 78.30 & 92.43 & 78.24 & 79.07 & 79.70 & 76.04 & 98.09 & 78.22 & 8.00 \\
GIN          & 88.38 & 81.97 & 73.80 & 74.17 & 76.61 & 75.58 & 96.22 & 91.98 & 9.50 \\
GraphSAGE    & 90.21 & \textcolor{rankthird}{\textbf{96.18}} & 81.09 & 80.29 & \textcolor{rankthird}{\textbf{85.97}} & \textcolor{rankthird}{\textbf{79.10}} & 99.77 & 98.46 & \textcolor{ranksecond}{\textbf{3.50}} \\
PC-GNN       & 86.68 & 92.28 & \textcolor{rankthird}{\textbf{84.04}} & \textcolor{rankthird}{\textbf{79.80}} & \textcolor{ranksecond}{\textbf{86.69}} & \textcolor{ranksecond}{\textbf{79.18}} & 99.43 & 98.25 & \textcolor{rankthird}{\textbf{4.75}} \\
BWGNN        & 84.63 & 84.08 & 75.10 & 73.74 & 79.44 & 71.31 & \textcolor{rankthird}{\textbf{99.78}} & \textcolor{rankthird}{\textbf{98.74}} & 7.75 \\
GHRN         & 74.53 & 78.60 & 74.85 & 73.39 & 77.64 & 70.27 & 99.68 & \textcolor{ranksecond}{\textbf{98.76}} & 9.12 \\
XGBGraph     & 88.72 & 92.83 & 79.34 & 78.66 & 85.59 & 77.64 & 99.56 & 96.49 & 5.38 \\
RFGraph      & 88.20 & 82.84 & 79.39 & 76.90 & 81.50 & 75.54 & 98.50 & 91.59 & 7.75 \\
GAAP         & 63.16 & 82.19 & 67.36 & 65.66 & 69.92 & 65.23 & 87.10 & 85.34 & 13.38 \\
PMP          & 61.21 & 84.37 & 66.83 & 66.63 & 72.69 & 60.79 & 86.52 & 82.83 & 13.62 \\
\midrule
\multicolumn{10}{l}{\textit{Integrating LLMs with graphs}} \\
TAPE         & \textcolor{ranksecond}{\textbf{91.87}} & \textcolor{ranksecond}{\textbf{98.30}} & \textcolor{ranksecond}{\textbf{87.38}} & \textcolor{ranksecond}{\textbf{88.84}} & 89.24 & 84.53 & 90.65 & 97.42 & 3.62 \\
CoLL         & 71.44 & 74.41 & 76.95 & 72.88 & 75.33 & 68.79 & 84.39 & 82.61 & 12.62 \\
UNPrompt     & 73.90 & 67.67 & 74.09 & 59.62 & 63.92 & 64.09 & 82.55 & 80.21 & 14.88 \\
GraphGPT     & 68.84 & 60.31 & 56.11 & 54.54 & 60.45 & 54.90 & 68.71 & 66.67 & 17.50 \\
InstructGLM  & 50.77 & 72.62 & 59.04 & 62.63 & 64.55 & 60.72 & 73.65 & 74.93 & 16.50 \\
ConsisGAD    & 63.32 & 77.89 & 66.93 & 67.24 & 70.01 & 58.42 & 86.59 & 69.59 & 14.75 \\
\midrule
\multicolumn{10}{l}{\textit{Ours}} \\
\rowcolor{gray!15}
\textbf{\textsc{\model}} & \textcolor{rankfirst}{\textbf{92.08}} & \textcolor{rankfirst}{\textbf{99.70}} & \textcolor{rankfirst}{\textbf{96.19}} & \textcolor{rankfirst}{\textbf{95.49}} & \textcolor{rankfirst}{\textbf{97.93}} & \textcolor{rankfirst}{\textbf{93.85}} & \textcolor{rankfirst}{\textbf{99.93}} & \textcolor{rankfirst}{\textbf{99.98}} & \textcolor{rankfirst}{\textbf{1.00}} \\
Improve.
  & +0.21 & +1.40 & +8.81 & +6.65 & +8.69 & +9.32 & +0.15 & +1.22 & -- \\
\bottomrule
\end{tabular}%
}
\caption{AUROC comparison across the eight benchmark datasets.
  \textcolor{rankfirst}{\textbf{First}},
  \textcolor{ranksecond}{\textbf{second}}, and
  \textcolor{rankthird}{\textbf{third}} best results are color-coded.}
\label{tab:auroc-full}
\end{table*}

\begin{table*}[!ht]
\centering
\footnotesize
\setlength{\tabcolsep}{4pt}
\resizebox{\textwidth}{!}{%
\begin{tabular}{lrrrrrrrrr}
\toprule
Method & Citeseer & Pubmed & History & Photo & Computers & Children & ogbn-Arxiv & CitationV8 & Avg.\ Rank \\
\midrule
\multicolumn{10}{l}{\textit{GNN-based methods}} \\
GCN          & \textcolor{ranksecond}{\textbf{67.37}} & 54.63 & 50.68 & 48.86 & 54.24 & 34.78 & 69.11 & 65.73 & 6.12 \\
GAT          & 27.49 & 50.24 & 42.09 & 45.66 & 52.11 & \textcolor{rankthird}{\textbf{39.00}} & 81.94 & 43.04 & 8.12 \\
GIN          & 62.39 & 45.16 & 49.87 & \textcolor{ranksecond}{\textbf{51.39}} & \textcolor{ranksecond}{\textbf{56.88}} & \textcolor{ranksecond}{\textbf{51.79}} & 75.55 & 68.21 & 5.88 \\
GraphSAGE    & \textcolor{rankthird}{\textbf{68.10}} & \textcolor{ranksecond}{\textbf{76.82}} & 45.38 & 38.46 & 50.76 & 33.77 & \textcolor{ranksecond}{\textbf{98.19}} & \textcolor{ranksecond}{\textbf{91.09}} & \textcolor{rankthird}{\textbf{4.75}} \\
PC-GNN       & 46.19 & 65.88 & 42.58 & 28.16 & 45.66 & 28.76 & \textcolor{rankthird}{\textbf{95.72}} & \textcolor{rankthird}{\textbf{88.48}} & 7.00 \\
BWGNN        & 61.62 & 46.49 & 16.28 & 15.93 & 29.44 & 13.34 & 98.41 & 91.11 & 8.25 \\
GHRN         & 39.81 & 40.64 & 17.50 & 16.37 & 27.83 & 13.49 & 97.58 & 90.84 & 9.38 \\
XGBGraph     & 56.30 & 61.74 & \textcolor{ranksecond}{\textbf{53.98}} & \textcolor{rankthird}{\textbf{45.69}} & \textcolor{rankthird}{\textbf{56.96}} & 35.51 & 96.33 & 78.94 & \textcolor{ranksecond}{\textbf{4.50}} \\
RFGraph      & 60.47 & 33.68 & \textcolor{rankthird}{\textbf{53.34}} & 41.32 & 52.34 & 29.92 & 86.46 & 66.28 & 7.38 \\
GAAP         & 12.30 & 51.73 & 11.68 & 11.16 & 20.03 & 9.13 & 75.73 & 61.65 & 12.25 \\
PMP          & 9.45 & 57.95 & 11.80 & 13.94 & 18.51 & 6.27 & 71.66 & 47.67 & 13.12 \\
\midrule
\multicolumn{10}{l}{\textit{Integrating LLMs with graphs}} \\
TAPE         & 63.26 & \textcolor{rankthird}{\textbf{83.61}} & 40.20 & 40.55 & 45.30 & 27.12 & 31.67 & 73.09 & 7.88 \\
CoLL         & 17.49 & 11.38 & 17.56 & 10.59 & 14.32 & 8.52 & 27.44 & 39.65 & 14.50 \\
UNPrompt     & 17.47 & 9.97 & 22.10 & 7.23 & 9.69 & 8.07 & 36.63 & 34.62 & 14.88 \\
GraphGPT     & 7.06 & 5.82 & 4.75 & 4.61 & 5.69 & 4.56 & 9.84 & 8.62 & 17.88 \\
InstructGLM  & 4.24 & 51.52 & 7.51 & 15.74 & 20.55 & 7.15 & 53.47 & 54.08 & 13.75 \\
ConsisGAD    & 9.93 & 42.74 & 9.90 & 14.31 & 20.89 & 4.83 & 71.38 & 15.11 & 14.38 \\
\midrule
\multicolumn{10}{l}{\textit{Ours}} \\
\rowcolor{gray!15}
\textbf{\textsc{\model}} & \textcolor{rankfirst}{\textbf{86.18}} & \textcolor{rankfirst}{\textbf{98.18}} & \textcolor{rankfirst}{\textbf{86.25}} & \textcolor{rankfirst}{\textbf{73.45}} & \textcolor{rankfirst}{\textbf{89.41}} & \textcolor{rankfirst}{\textbf{71.53}} & \textcolor{rankfirst}{\textbf{99.71}} & \textcolor{rankfirst}{\textbf{99.78}} & \textcolor{rankfirst}{\textbf{1.00}} \\
Improve.
  & +18.08 & +14.57 & +32.27 & +22.06 & +32.45 & +19.74 & +1.30 & +8.67 & -- \\
\bottomrule
\end{tabular}%
}
\caption{AUPRC comparison across the eight benchmark datasets.
  \textcolor{rankfirst}{\textbf{First}},
  \textcolor{ranksecond}{\textbf{second}}, and
  \textcolor{rankthird}{\textbf{third}} best results are color-coded.}
\label{tab:auprc-full}
\end{table*}

\section{NodeScaler Analysis}
\label{app:nodescaler-full}

The three-panel NodeScaler visualization
(Figure~\ref{fig:nodescaler} in the main body) provides mechanistic
evidence that NCM learns meaningful, label-discriminative modulation
rather than collapsing to the standard parameter update.
Panel~(a) plots mean $|s_j{-}1|$ per channel averaged over the
top-5 most discriminative layers. Normal nodes exhibit consistently
higher deviation than anomaly nodes across all 64 channels,
indicating that NCM selectively amplifies channel activations for
normal nodes while leaving anomaly nodes closer to the default
behavior.
Panel~(b) shows the per-block discriminative power of NodeScaler
deviations. The majority of mid-to-late Transformer blocks are strongly
discriminative, with normal nodes receiving larger modulation, while a
small number of early blocks show the opposite direction, revealing a
structured, depth-dependent modulation pattern.
Panel~(c) presents the violin plot of per-node mean $|s{-}1|$,
confirming that normal nodes receive systematically stronger modulation
than anomaly nodes.

\section{Qualitative Case Study}
\label{app:qualitative}

We present a real case from Pubmed illustrating how \textsc{\model}
detects an anomaly that the text-only variant misses.

\paragraph{Setup.}
Node 7229 is a structure-perturbed anomaly (degree 6, avg.\ neighbor
similarity $0.30$). Its text is a meta-analysis of the Gly482Ser
variant in PPARGC1A in type 2 diabetes. The two most similar neighbors
(similarity $0.51$ and $0.47$) discuss closely related PPARGC1A
genetics. However, the two most dissimilar neighbors (similarity
$-0.10$ and $-0.08$) discuss cerebral blood flow measurement and
herbal extracts for dyslipidemia, topics unrelated to the target node's
genetics focus.

\paragraph{Text-only prediction.}
Without neighborhood context, the LLM assigns a near-zero anomaly score
($0.0032$) and predicts \textit{normal}, because the abstract is
well-formed and topically coherent in isolation.

\begin{table*}[!ht]
\centering
\footnotesize
\setlength{\tabcolsep}{4pt}
\resizebox{\textwidth}{!}{%
\begin{tabular}{lcccccccc}
\toprule
Variant & Citeseer & Pubmed & History & Photo & Computers & Children & ogbn-Arxiv & CitationV8 \\
\midrule
Text        & 56.41 & 65.25 & 52.08 & 59.22 & 57.00 & 48.79 & 65.34 & 65.89 \\
Text+NCM    & 56.41 & 65.25 & 51.64 & 58.82 & 55.78 & 64.95 & 93.35 & 94.38 \\
Stats       & 89.00 & 82.06 & 70.31 & 59.29 & 62.15 & 52.14 & 83.90 & 87.89 \\
Stats+NCM   & 89.77 & 84.41 & 71.53 & 59.29 & 82.54 & 65.65 & 95.81 & 97.00 \\
Explicit    & 86.54 & 92.83 & 83.71 & 65.49 & 72.25 & 68.10 & 98.22 & 98.41 \\
\rowcolor{gray!15}
\textbf{\textsc{\model}} & \textbf{87.50} & \textbf{93.46} & \textbf{84.70} & \textbf{66.76} & \textbf{85.78} & \textbf{70.40} & \textbf{98.35} & \textbf{98.53} \\
\midrule
SimOnly+NCM    & 89.80 & 85.62 & 71.19 & 56.34 & 83.89 & 65.39 & 96.21 & 96.84 \\
DissimOnly+NCM & 84.62 & 93.38 & 80.59 & 66.38 & 83.22 & 68.74 & 98.64 & 98.74 \\
Random+NCM     & 88.00 & 88.10 & 79.20 & 72.12 & 86.41 & 64.63 & 96.28 & 97.81 \\
Shuffled+NCM   & 89.80 & 83.83 & 69.10 & 57.24 & 82.23 & 63.48 & 95.89 & 97.06 \\
\bottomrule
\end{tabular}%
}
\caption{Full ablation results (Anomaly-F1). Text = text-only input.
  Stats = graph-structural neighborhood semantics only. Explicit = full
  explicit path without NCM. \textsc{\model} = full system (explicit + NCM). +NCM
  indicates Neighborhood Context Modulation is active.}
\label{tab:ablation-full}
\end{table*}

\paragraph{\textsc{\model} prediction.}
With the explicit fusion path, the LLM receives both similar and
dissimilar neighbor texts alongside graph statistics. The low average
neighbor similarity ($0.30$) combined with the semantically distant
dissimilar neighbors enables the model to identify that this node's
PPARGC1A genetics content is inconsistent with its broader neighborhood
context. The model assigns a high anomaly score ($0.9844$) and correctly
predicts \textit{anomaly}.

\begin{table*}[h]
\centering
\footnotesize
\setlength{\tabcolsep}{4pt}
\resizebox{\textwidth}{!}{%
\begin{tabular}{lcccccccc}
\toprule
Variant & Citeseer & Pubmed & History & Photo & Computers & Children & ogbn-Arxiv & CitationV8 \\
\midrule
Text        & 70.24 & 74.68 & 71.21 & 75.27 & 71.63 & 70.50 & 74.09 & 74.87 \\
Text+NCM    & 69.69 & 77.34 & 71.26 & 75.59 & 72.14 & 90.27 & 99.33 & 99.60 \\
Stats       & 91.79 & 98.04 & 94.21 & 91.36 & 94.08 & 84.16 & 99.07 & 99.58 \\
Stats+NCM   & 92.77 & 98.23 & 94.15 & 90.68 & 97.28 & 91.54 & 99.79 & 99.91 \\
Explicit    & 91.98 & 99.68 & 96.12 & 95.32 & 96.82 & 93.05 & 99.88 & 99.97 \\
\rowcolor{gray!15}
\textbf{\textsc{\model}} & \textbf{92.08} & \textbf{99.70} & \textbf{96.19} & \textbf{95.49} & \textbf{97.93} & \textbf{93.85} & \textbf{99.93} & \textbf{99.98} \\
\midrule
SimOnly+NCM    & 92.27 & 98.49 & 94.22 & 91.27 & 97.25 & 91.69 & 99.82 & 99.95 \\
DissimOnly+NCM & 91.23 & 99.41 & 96.15 & 95.24 & 97.98 & 93.67 & 99.91 & 99.98 \\
Random+NCM     & 90.96 & 98.74 & 95.12 & 95.94 & 98.01 & 91.66 & 99.85 & 99.97 \\
Shuffled+NCM   & 91.09 & 98.11 & 92.98 & 90.71 & 97.66 & 91.05 & 99.76 & 99.93 \\
\bottomrule
\end{tabular}%
}
\caption{Full ablation results (AUROC). Same variant naming as
  Table~\ref{tab:ablation-full}.}
\label{tab:ablation-auroc}
\end{table*}

\begin{table*}[h]
\centering
\footnotesize
\setlength{\tabcolsep}{4pt}
\resizebox{\textwidth}{!}{%
\begin{tabular}{lcccccccc}
\toprule
Variant & Citeseer & Pubmed & History & Photo & Computers & Children & ogbn-Arxiv & CitationV8 \\
\midrule
Text        & 44.62 & 52.94 & 41.87 & 49.16 & 46.53 & 39.88 & 53.19 & 53.99 \\
Text+NCM    & 44.57 & 53.40 & 42.65 & 48.80 & 46.06 & 65.35 & 97.05 & 97.45 \\
Stats       & 85.89 & 86.46 & 72.77 & 60.18 & 66.03 & 50.56 & 90.65 & 94.46 \\
Stats+NCM   & 86.34 & 87.67 & 73.10 & 58.98 & 86.32 & 67.12 & 98.45 & 99.15 \\
Explicit    & 85.84 & 97.79 & 85.42 & 73.43 & 78.67 & 71.38 & 99.67 & 99.71 \\
\rowcolor{gray!15}
\textbf{\textsc{\model}} & \textbf{86.18} & \textbf{98.18} & \textbf{86.25} & \textbf{73.45} & \textbf{89.41} & \textbf{71.53} & \textbf{99.71} & \textbf{99.78} \\
\midrule
SimOnly+NCM    & 84.69 & 89.48 & 73.26 & 60.09 & 87.35 & 66.72 & 98.61 & 99.25 \\
DissimOnly+NCM & 80.19 & 97.22 & 83.58 & 72.23 & 87.41 & 69.97 & 99.74 & 99.84 \\
Random+NCM     & 84.99 & 93.29 & 81.33 & 77.57 & 88.61 & 67.30 & 99.00 & 99.52 \\
Shuffled+NCM   & 86.28 & 87.51 & 70.42 & 58.09 & 86.87 & 64.15 & 98.48 & 99.28 \\
\bottomrule
\end{tabular}%
}
\caption{Full ablation results (AUPRC). Same variant naming as
  Table~\ref{tab:ablation-full}.}
\label{tab:ablation-auprc}
\end{table*}

This case demonstrates that anomalies arising from neighborhood
semantic inconsistency require
neighborhood-level semantic reasoning that is unavailable to methods
operating on node text alone.

\subsection{Explicit Fusion Path Ablation (Full Results)}

Table~\ref{tab:ablation-full} reports Anomaly-F1 for all ablation
variants across the eight datasets. The core variants (Text,
Text+NCM, Stats, Stats+NCM, Explicit, and \textsc{\model}) and the
contrastive-neighbor controls (SimOnly+NCM, DissimOnly+NCM,
Random+NCM, Shuffled+NCM) all have results on the full set of eight
datasets.

\section{The Use of Large Language Models}
\label{app:llm-use}

In this paper, large language models were utilized exclusively for
grammatical polishing and stylistic refinement, aimed at enhancing the
clarity and readability of our presentation of results and conclusions.

\end{document}